\documentclass{article}

\usepackage[preprint]{corl_2022} 

\usepackage{multicol}
\usepackage{graphicx}

\usepackage{balance}
\usepackage{hyperref}
\hypersetup{
    colorlinks=true,
    linkcolor=blue,
    filecolor=magenta,      
    urlcolor=cyan,
    citecolor=blue,
}
\usepackage{times}
\usepackage{amsfonts}
\usepackage[ruled,vlined]{algorithm2e}
\usepackage{amsmath}
\usepackage{wrapfig}
\usepackage{lipsum, booktabs}
\usepackage{enumerate}
\newcommand{\MARL}{MARL}
\newcommand{\rl}{reinforcement learning}

\newtheorem{proposition}{Proposition}
\newtheorem{definition}{Definition}

\title{Stackelberg Games for Learning Emergent Behaviors During Competitive Autocurricula}

\author{
  Boling~Yang\textsuperscript{\rm 1},
  Liyuan~zheng\textsuperscript{\rm 1},
  Lillian J. Ratliff\textsuperscript{\rm 1},
  Byron Boots\textsuperscript{\rm 1},
  Joshua R.~Smith\textsuperscript{\rm 1}\\
  \textsuperscript{\rm 1} The University of Washington
}

\begin{document}
\maketitle


\begin{abstract}
    Autocurricula training is an important sub-area of multi-agent reinforcement learning~(MARL). In the robotics community, autocurricula has experimented with physically grounded problems, such as robust control and interactive manipulation tasks. However, the asymmetric nature of these tasks makes the generation of sophisticated policies challenging. Indeed, the asymmetry in the environment may implicitly or explicitly provide an advantage to a subset of agents which could, in turn, lead to a low-quality equilibrium. This paper proposes a novel game-theoretic MARL algorithm, Stackelberg Multi-Agent Deep Deterministic Policy Gradient (ST-MADDPG), which formulates a two-player MARL problem as a Stackelberg game with one player as the `leader' and the other as the `follower' in a hierarchical interaction structure wherein the leader has an advantage. In three asymmetric competitive robotics environments, we exploit the leader's advantage from ST-MADDPG to improve the quality of autocurricula training and result in more sophisticated and complex autonomous agents.
    
\end{abstract}


\section{Introduction}
\label{sec:intro}
Multi-agent Reinforcement Learning (MARL) addresses the sequential decision-making problem of multiple autonomous agents that interact with each other in a common environment, each of which aims to optimize its own long-term return~\citep{zhang2019multi}.  Purely competitive settings form an important class of sub-problems in {\MARL}, and are typically formulated as a zero-sum two-player game using the framework of competitive Markov decision processes~\citep{filar2012competitive}. There has been much success in using competitive {\MARL} methods to solve such problems, especially for symmetric games including extensive form games on finite action spaces such as chess and video games~\citep{silver2017mastering, berner2019dota, vinyals2019grandmaster}. These algorithms typically use a co-evolution training scheme in which the competing agents continually create new tasks for each other and incrementally improve their own policies by solving these new tasks. However, once one or more evolved agents fails to sufficiently challenge their opponent, subsequent training is unlikely to result in further progress due to a lack of pressure for adaptation. This cessation of the co-evolution process indicates that the agents have reached an equilibrium.


Recently, competitive {\MARL} methods have gained attention from the robotics community and have been used to solve physically grounded problems, such as adversarial learning for robust control, autonomous task generation, and complex robot behavior learning~\citep{dennis2020emergent,baker2019emergent,yang2021motivating}. However, these problems are typically asymmetric in practice. Unlike a symmetric game where all agents have the same knowledge and the same ability to act, an asymmetric game requires the agents to solve their own task while coupled in an imbalanced competitive environment.  One agent could gain advantage from having an easier initial task, and learn to exploit the advantage to quickly dominate the game~\citep{pinto2017robust}. This will prematurely terminate the co-evolving process and all agents are trapped in a low quality equilibrium. For example, in a simulated boxing game, if a player is able to punch significantly harder than the other, it can easily execute a knockout. Such a player could learn to knock out the opponent at the very beginning of a match, leaving no chance for the opponent to explore for better counter strategies such as strategic footwork to avoid the knockout blow.

To overcome this challenge, one common approach is to simply generate a large amount of diversified samples using population-based methods and distributed sampling~\citep{bansal2017emergent, baker2019emergent}. Yet, this approach only partially mitigate the issue. With a sufficient amount of engineering effort, some policy initialization methods such as reward shaping and imitation learning could be used to initialize the system to a desired state~\citep{won2021control, yang2021motivating}. The minimax regret strategy is a risk-neutral decision-maker that has been demonstrated to prevent the stronger player from dominating the game in adversarial learning~\citep{dennis2020emergent}. More importantly, by simply treating two players equally with a symmetric information structure and simultaneous learning dynamics, these methods fail to capture the inherent imbalanced underlying structure of the environment. In addition, as presented in~\citep{foerster2017learning, prajapat2020competitive, zheng2021stackelberg}, MARL methods that use simultaneous gradient descent ascent updates could result in poor convergence properties in practice.
\begin{figure*}[t!]
\centering
\includegraphics[width=0.90\textwidth]{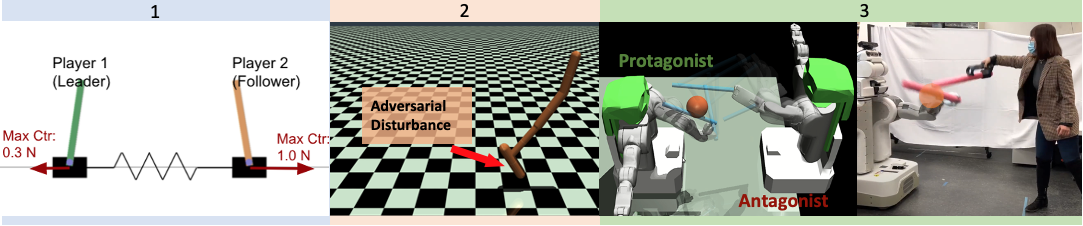}
\caption{This work focuses on three competitive robotics tasks with physical interaction. 1.~\textbf{Competitive-cartpoles} is a simple one-dimensional continuous control environment. 2.~\textbf{Hopper with adversarial disturbances} is a classic robust control problem. 3.~\textbf{The fencing game} is a competitive-HRI benchmark. Video demonstration of the robots' behaviors in various environments can be found in \href{https://sites.google.com/view/stackelberg-autocurricula/}{the project website}.}
\label{fig:all_env}
\end{figure*}

We aim to solve this problem by directly modifying the game dynamics to aid in re-balancing the bias in asymmetric environments. 
Asymmetric gradient-based algorithms in multi-agent learning have been actively studied in recent years in normal form game settings~\citep{lin2020near, fiez2020implicit}.
In this paper, we leverage the Stackelberg game structure~\citep{von2010market} to introduce a hierarchical order of play, and therefore an asymmetric interaction structure, into competitive Markov games~\citep{yang2020overview}. 

In a two-player Stackelberg game, the leader knows that the follower will react to its announced strategy. 
As a result of this structure, the leader optimizes its objective accounting for the anticipated response of the follower, while the follower selects a myopic best response to the leader's action to optimize its own objective. As a result, the leader stands to benefit from the Stackelberg game structure by achieving a better equilibrium payoff compared to that in a normal competitive game~\citep{bacsar1998dynamic}. This is a desirable property when one agent is the primary agent in the task (e.g., robust control with adversaries) or when one agent has initial or inherent disadvantage due to the asymmetric game environment and a re-balance of power is sought, as we will demonstrate in our experiments. The main \textbf{contributions} of this paper are listed below:

\paragraph{A Novel MARL Algorithm: ST-MADDPG.} We formulate the two-plater competitive MARL problem as a Stackelberg game. By adopting the total derivative Stackelberg learning update rule, we extend the current state-of-the-art MARL algorithm MADDPG~\citep{lowe2017multi} to a novel Stackelberg version, termed Stackelberg MADDPG (ST-MADDPG).

\paragraph{Better Competitive Learning in Asymmetric Environment.} From a novel perspective, we study how information and force exertion asymmetries affect the agents' performance and behaviors during the multi-agent co-evolution process. We first design a simple competitive RL benchmark with continuous control space: \emph{competitive-cartpoles} (Figure~\ref{fig:all_env}.1). In this environment, we demonstrate that the use of the Stackelberg gradient updates provide an advantage to the leader agent that compensates for the agent's initial or inherent disadvantage and leads to better performance.

\paragraph{Application to Practical Robotics Tasks.}The practical effectiveness of the proposed algorithm is demonstrated in two tasks. In a robust control problem (Figure~\ref{fig:all_env}.2), having the leader advantage during adversarial training allows the resulting robot to better survive adversarial and intense random disturbances. In a competitive fencing game (Figure~\ref{fig:all_env}.3), ST-MADDPG allows some attacking agents to learn complex strategies for better performance. Notably, two of the best performing attackers learned to trick the opponent to move to less manipulable joint configurations which temporary limit the opponent's maneuverability.

\section{Preliminaries}
In this section, we provide the requisite preliminary mathematical model and notation.
\paragraph{Two-player Competitive Markov Game.}
We consider a two-player zero-sum fully observable competitive Markov game (i.e., competitive MDP). A competitive Markov game is a tuple of $(\mathcal{S}, \mathcal{A}^1, \mathcal{A}^2, P, r)$, where $\mathcal{S}$ is the state space, $s \in \mathcal{S}$ is a state, player $i \in \{ 1, 2\}$, $A^i$ is the player $i$'s action space with $a^i \in \mathcal{A}^i$. $P: \mathcal{S} \times \mathcal{A}^1 \times \mathcal{A}^2 \rightarrow \mathcal{S}$ is the transition kernel such that $P(s'|s, a^1, a^2)$ is the probability of transitioning to state $s'$ given that the previous state was $s$ and the agents took action $(a^1, a^2)$ simultaneously in $s$. Reward $r: \mathcal{S} \times \mathcal{A}^1 \times \mathcal{A}^2 \rightarrow \mathbb{R}$ is the reward function of player 1 and by the zero-sum nature of the competitive setting, player 2 receives the negation of $r$ as its own reward feedback. Each agent uses a stochastic policy $\pi_{\theta}^i$, parameterized by $\theta^i$. 

A trajectory $\tau = (s_0, a_0^1, a_0^2, \dots, s_T, a_T^1, a_T^2)$ gives the cumulative rewards or return defined as $R(\tau) =  \sum_{t=0}^T \gamma^t r(s_t, a_t^1, a_t^2)$, where the discount factor $0 < \gamma \le 1$ assigns weights to rewards received at different time steps. The expected return of $\pi = \{\pi^1, \pi^2\}$ after executing joint action profile $(a_t^1, a_t^2)$ in state $s_t$ can be expressed by the following $Q^\pi$ function: 
\begin{gather}
\textstyle	Q^\pi(s_t, a_t^1, a_t^2) = \mathbb{E}_{\tau \sim \pi} \big[ \sum_{t' = t}^T \gamma^{t'-t} r(s_{t'}, a_{t'}^1, a_{t'}^2) | s_t, a_t^1, a_t^2 \big],\notag 
\end{gather}
where $\tau \sim \pi$ is shorthand to indicate that the distribution over trajectories depends on $\pi: s_0 \sim \rho, a^1_t \sim \pi^1(\cdot|s_t), a^2_t \sim \pi^2(\cdot|s_t), s_{t+1} \sim P(\cdot | s_t, a^1_t, a^2_t)$. $\rho$ is the system initial state distribution.

The game objective is the expected return and is given by
\begin{equation}
J(\pi) = \textstyle \mathbb{E}_{\tau \sim \pi}  \big[ \sum_{t = 0}^T \gamma^{t} r(s_t, a_t^1, a_t^2) \big]  = \textstyle \mathbb{E}_{s \sim \rho, a^1 \sim \pi^1(\cdot|s), a^2 \sim \pi^2(\cdot|s)} \big[ Q^\pi(s, a^1, a^2) \big].\notag
\end{equation}
In a competitive Markov game, player 1 aims to find a policy maximizing the game objective, while player 2 aims to minimize it. That is, they solve for $\max_{\theta_1} J(\pi^1, \pi^2)$ and $\min_{\theta_2} J(\pi^1, \pi^2)$, respectively.

\paragraph{Stackelberg Game Preliminaries.}
\label{sec:stackgame}
A Stackelberg game is a game  between two agents where one agent is deemed the leader and the other the follower.  Each agent has an objective they want to optimize that depends on not only their own actions but also the actions of the other agent. Specifically, the leader optimizes its objective under the assumption that the follower will play a best response. Let $J_1(\theta_1, \theta_2)$ and $J_2(\theta_1, \theta_2)$ be the objective functions that the leader and follower want to minimize (in a competitive setting $J_2 = -J_1$), respectively, where $\theta_1\in \Theta_1\subseteq\mathbb{R}^{d_1}$ and $\theta_2\in \Theta_2\subseteq\mathbb{R}^{d_2}$ are their decision variables or strategies and $\theta=(\theta_1,\theta_2)\in \Theta_1\times \Theta_2$ is their joint strategy. 
The leader and follower aim to solve the following problems:
\begin{align}
&\textstyle \max_{\theta_1\in \Theta_1} \ \{ J_1(\theta_1,\theta_2)\ \big|\ \theta_2\in \arg\max_{\theta_2\in \Theta_2}J_2(\theta_1,\theta_2)\}, \tag{ L}\\
&\textstyle \max_{\theta_2\in \Theta_2} \ J_2(\theta_1,\theta_2). \tag{ F}
\end{align}

Since the leader assumes the follower chooses a best response $\theta_2^{*}(\theta_1) = \arg\max_{\phi} J_2 (\theta_1, \theta_2)$,
the follower’s decision variables are implicitly a function of the leader’s. In deriving sufficient conditions for the optimization problem in (L), the leader utilizes this information in computing the total derivative of its cost:
\begin{equation*}
\nabla J_1(\theta_1, \theta_2^*(\theta_1)) = \nabla_{\theta_1}J_1(\theta) + (\nabla \theta_2^*(\theta_1))^\top\nabla_{\theta_2}J_1(\theta),
\label{eq:stac_grad}
\end{equation*}
where
$\nabla \theta_2^*(\theta_1) = - ( \nabla_{\theta_2}^2J_2(\theta))^{-1}\nabla_{\theta_2 \theta_1}J_2(\theta)$\footnote{The partial derivative of $J(\theta_1,\theta_2)$ with respect to the $\theta_i$ is denoted by $\nabla_{\theta_i}J(\theta_1,\theta_2)$ and the total derivative of $J(\theta_1,h(\theta_1))$ for some function $h$, is denoted $\nabla J$ where $\nabla J(\theta_1,h(\theta_1)=\nabla_{\theta_1}J(\theta_1,h(\theta_1))+(\nabla h(\theta_1))^\top \nabla_{\theta_2}J(\theta_1,h(\theta_1))$.}
by the implicit function theorem~\citep{krantz2002implicit}.

A point $\theta=(\theta_1,\theta_2)$ is a local solution to (L) if $\nabla J_1(\theta_1,\theta_2^\ast(\theta_1))=0$ and $\nabla^2 J_1(\theta_1,\theta_2^\ast(\theta_1))>0$. For the follower's problem, sufficient conditions for optimality are $\nabla_{\theta_2}J_2(\theta_1,\theta_2)=0$ and $\nabla_{\theta_2}^2J_2(\theta_1,\theta_2)>0$. This gives rise to the following equilibrium concept which characterizes sufficient conditions for a local Stackelberg equilibrium.

\begin{definition}[Differential Stackelberg Equilibrium,~\citealt{fiez2020implicit}]
	The joint strategy profile $\theta^{\ast}$ $=(\theta_1^\ast,\theta_2^\ast)\in \Theta_1\times \Theta_2$ is a differential Stackelberg equilibrium if $\nabla J_1(\theta^\ast)=0$, $\nabla_{\theta_2}J_2(\theta^\ast)=0$, $\nabla^2J_1(\theta^\ast)>0$, and $\nabla_{\theta_2}^2J_2(\theta^\ast)>0$.
	\label{def:stackelberg}
\end{definition}

The Stackelberg learning dynamics derive from the first-order gradient-based sufficient conditions and are given by 
$\theta_{1,k+1}=\theta_{1,k}-\alpha_1 \nabla J_1(\theta_{1,k},\theta_{2,k})$, and
$\theta_{2,k+1}=\theta_{2,k}-\alpha_2 \nabla_{\theta_2} J_2(\theta_{1,k},\theta_{2,k})$,
where $\alpha_i$, $i=1,2$ are the leader and follower learning rates.

\paragraph{MADDPG.}
\citet{lowe2017multi} showed that na\"ive policy gradient methods perform poorly in simple multi-agent continuous control tasks and proposed a more advanced MARL algorithm termed MADDPG, which is one of the state-of-the-art multi-agent control algorithms. The idea of MADDPG is to adopt the framework of centralized training with decentralized execution. Specifically, they use a centralized critic network $Q_w$ to approximate the $Q^\pi$ function, and update the policy network $\pi^i_\theta$ of each agent using the global critic. Consider the deterministic policy setting where each player has policy $\mu_{\theta_i}$ with parameter $\theta_i$.\footnote{Following the setting and notation in origin DDPG algorithm~\citep{lowe2017multi}, we use $\mu$ to represent deterministic policy to differentiate it from stochastic ones.} The game objective (for player 1) is $J(\theta_1, \theta_2) = \mathbb{E}_{\xi \sim \mathcal{D}} \big[ Q_w(s, \mu_{\theta_1}(s), \mu_{\theta_2}(s)) \big]$, where $\xi=(s,a^1, a^2,r,s')$, $\mathcal{D}$ is a replay buffer. The policy gradient of each player can be computed as
$\nabla_{\theta_1} J(\theta_1, \theta_2) = \mathbb{E}_{\xi \sim \mathcal{D}} \big[ \nabla_{\theta_1} \mu_{\theta_1}(s) \nabla_{a^1} Q_w(s, a^1, a^2) |_{a^1 = \mu_{\theta_1}(s)} \big],$ and $ \nabla_{\theta_2} J(\theta_1, \theta_2) = \mathbb{E}_{\xi \sim \mathcal{D}} \big[ \nabla_{\theta_2} \mu_{\theta_2}(s) \nabla_{a^2} Q_w(s, a^1, a^2) |_{a^2 = \mu_{\theta_2}(s)} \big]$.

The critic objective is defined as the mean square Bellman error
$L(w) = \mathbb{E}_{\xi \sim \mathcal{D}} [ (Q_w(s, a^1, a^2) - (r + \gamma Q_{w'}(s', \mu_{\theta_1'}(s'), \mu_{\theta_2'}(s')) )^2 ]$,
where $Q_{w'}$ and $\mu_{\theta_1'}, \mu_{\theta_2'}$ are target networks obtained by polyak averaging the $Q_w$ and $\mu_{\theta_1}, \mu_{\theta_2}$ network parameters over the course of training.

With MADDPG in competitive settings, the centralized critic is updated by gradient descent and the two agents' policies are updated by simultaneous gradient ascent and descent
$\theta_1 \leftarrow \theta_1 + \alpha^1 \nabla_{\theta_1} J(\theta_1, \theta_2)$, 
$\theta_2 \leftarrow \theta_2 - \alpha^2 \nabla_{\theta_2} J(\theta_1, \theta_2)$.

\section{Stackelberg MADDPG Algorithm}
\label{sec:stmaddpg}
In this section, we introduce our novel ST-MADDPG algorithm. A central feature of ST-MADDPG is that the leader agent exploits the knowledge that the follower will respond to its action in deriving its gradient based update. Namely, the total derivative learning update gives the advantage to the leader by anticipating the follower's update during learning and leads to Stackelberg equilibrium convergence in a wide range of applications such as generative adversarial networks and actor-critic networks~\citep{fiez2020implicit,zheng2021stackelberg}. According to~\citet[Chapter 4]{bacsar1998dynamic}, in the two-player game with unique follower best responses, the payoff of the leader in Stackelberg equilibrium is better than Nash equilibrium, which is desired in many applications. The full ST-MADDPG algorithm is shown in Algorithm~\ref{comp:alg:stmaddpg} in Appendix~\ref{sec:apdx_stmaddpg}.

Setting player 1 to be the leader, the ST-MADDPG policy gradient update rules for both players are given by:
\begin{align*}
    \theta_1 &\leftarrow \theta_1 + \alpha^1 \nabla J(\theta_1, \theta_2),\\
    \theta_2 &\leftarrow \theta_2 - \alpha^2 \nabla_{\theta_2} J(\theta_1, \theta_2),
\end{align*}
where the total derivative in the leader's update is given by
\begin{gather}
\nabla J(\theta_1, \theta_2) = \nabla_{\theta_1} J(\theta_1, \theta_2) -
\nabla_{\theta_1 \theta_2} J(\theta_1, \theta_2)(\nabla_{\theta_2}^2 J(\theta_1, \theta_2))^{-1} \nabla_{\theta_2} J(\theta_1, \theta_2). \label{comp:eqn:total_deri}
\end{gather}
The second order terms of the total derivative in~\eqref{comp:eqn:total_deri} can be computed by applying the chain rule:
\begin{align*}
    \nabla_{\theta_1 \theta_2} J(\theta_1, \theta_2) &= \mathbb{E}_{\xi \sim \mathcal{D}} [ \nabla_{\theta_1 } \mu_{\theta_1}(s) \nabla_{a^1 a^2} Q_w(s, a^1, a^2) (\nabla_{\theta_2 } \mu_{\theta_2}(s) )^T  |_{a^1 = \mu_{\theta_1}(s), a^2 = \mu_{\theta_2}(s)} ],\\
    \nabla_{\theta_2}^2 J(\theta_1, \theta_2) &= \mathbb{E}_{\xi \sim \mathcal{D}} \left[ \nabla_{\theta_2}^2 \mu_{\theta_2}(s) \nabla_{a^2} Q_w(s, a^1, a^2) |_{a^2 = \mu_{\theta_2}(s)} \right].
\end{align*}
To obtain an estimator of the total derivative $\nabla J(\theta_1, \theta_2)$, each part of~\eqref{comp:eqn:total_deri} is computed by sampling from a replay buffer. The inverse-Hessian-vector product can be efficiently computed by conjugate gradient~\citep{zheng2021stackelberg}. 

\paragraph{Implicit Map Regularization.} \label{sec:regularization}

The total derivative in the Stackelberg gradient dynamics requires computing the inverse of follower Hessian $\nabla_{\theta_2}^2 J(\theta_1, \theta_2)$. 
Since policy networks in practical {\rl} problems may be highly non-convex, $(\nabla_{\theta_2}^2 J(\theta_1, \theta_2))^{-1}$ can be ill-conditioned. 
Thus, instead of computing this term directly, in practice we compute a regularized variant of the form $(\nabla_{\theta_2}^2 J(\theta_1, \theta_2) + \lambda I)^{-1}$. This regularization method can be interpreted as the leader viewing the follower as optimizing a regularized cost $J(\theta_1, \theta_2)+\tfrac{\lambda}{2}\| \theta_2\|^2$,  while the follower actually optimizes $J(\theta_1, \theta_2)$.
The regularization $\lambda$ interpolates between the Stackelberg and individual gradient updates for the leader.
\begin{proposition}
	Consider a Stackelberg game where the leader updates using the regularized total gradient $\nabla^{\lambda} J_1(\theta) = \nabla_{\theta_1}J_1(\theta) - \nabla_{\theta_2 \theta_1}^{\top}J_2(\theta) (\nabla_{\theta_2}^2J_2(\theta) + \lambda I)^{-1} \nabla_{\theta_2}J_1(\theta)$.
	The following limiting conditions hold: 1) $\nabla^{\lambda}J_1(\theta)\rightarrow \nabla J_1(\theta)$ as $\lambda \rightarrow 0$; 2) $\nabla^{\lambda}J_1(\theta)\rightarrow \nabla_{\theta_1} J_1(\theta)$ as $\lambda \rightarrow \infty$.
\end{proposition}

\section{Experiments}
In this section, we report on three experiment environments that provide insight into the following questions:
\textbf{(Q1)}: How do different asymmetries in the training environment affect the performance and behavior of the agents? \textbf{(Q2)}: Can the leader's advantage from ST-MADDPG compensate for a weaker agent's inherent disadvantage? \textbf{(Q3)}: Is the total derivative update method used by ST-MADDPG better than an alternative approximation method? \textbf{(Q4)}: How to solve real-world robotics problems by exploiting the Stackelberg information structure?

Note that the trend of the cumulative reward of learning does not increase monotonically in most of the competitive MARL environments as in well trained single-agent or cooperative MARL environments. Hence, to evaluate an agent's performance, we choose to collect gameplay data by having the trained agents play multiple games against its co-evolving partner from training or a hand-designed reference opponent. Further execution details are described in Section~\ref{sec:learn_asym}, \ref{sec:hopper_adv} and \ref{sec:fencing}.

\begin{figure*}[t!]
\centering
\includegraphics[width=0.95\textwidth]{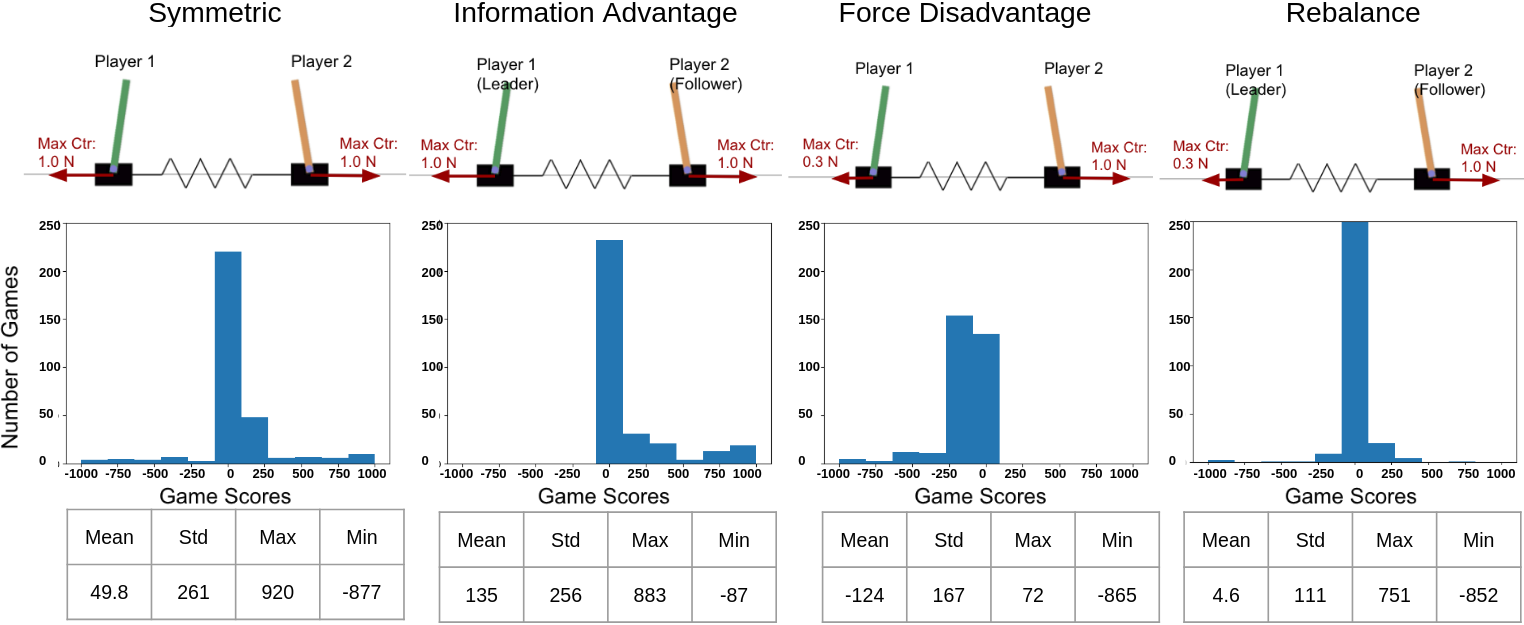}
\caption{Statistical analysis of the learned policies' performance in four different variations of the competitive-cartpoles environment. The game scores refer to Player~1's scores.}
\label{fig:c_cartpole}
\end{figure*}

\paragraph{Competitive-Cartpoles.} In order to answer \textbf{Q1} and \textbf{Q2}, we proposed a two-player zero-sum competitive game in which each agent solves a one-dimensional control task. As shown in Figure~\ref{fig:c_cartpole}, this environment contains two regular {cartpole} agents. The dynamics of the two agents are coupled by a spring, where each end of the spring connects to one of the agent's bodies. Both agents will get zero reward when they balance their own poles at the upright position simultaneously. If one of the agents loses its balance, this agent will receive a reward of $-1$ for every subsequent time step in the future until the game ends. The still balanced agent will get a reward of $+1$ for every time step until it also loses its balance and ends the game. As a result, the goal of each agent is to prevent its own pole from falling over, while seeking to break the balance of the opponent by introducing disturbing forces via the spring.

\paragraph{Hopper with Adversarial Disturbance.}   To investigate \textbf{Q3}, we will first focus on creating a robust control policy for the classic hopper environment using adversarial training~\citep{duan2016benchmarking, pinto2017robust}. Here, the first agent controls the classic hopper robot with four rigid links and three actuated joints. The second agent learns to introduce adversarial two-dimensional forces applied to the foot of the hopper.

\paragraph{The Fencing Game.}
To further examine \textbf{Q3}, we consider a zero-sum competitive game proposed by \citet{yang2021motivating}. This game represents more practical robotics challenges such as complex robot kinematics, high uncertainty transition dynamics, and highly asymmetric game mechanism. This game is a two player attack and defend game where the attacker aims to maximize its game score by attacking a predefined target area with a sword, without making contact with the protector's sword. The protector aims to minimize the attacker's score by defending the target area. The game rules are detailed in Appendix~\ref{sec:apdx_fencinggame}.

\subsection{Learning Under Asymmetric Advantage}
\label{sec:learn_asym}
This experiment explicitly studies the performance and behavior of the trained agents in the competitive-cartpoles environment under symmetric and asymmetric settings. We first demonstrate how ST-MADDPG can provide an advantage to an agent and improve its performance. We then show that given an asymmetric environment where one agent has a force exertion advantage over the other, ST-MADDPG can be used to retain a balance in agents' performance.

\paragraph{Leader Advantage.}
This experiment starts with a symmetric competitive-cartpoles environment, where both agents have the same ability to act. To understand how the leader advantage inherent in the Stackelberg game structure affects the auto-curriculum process, we ran both MADDPG and ST-MADDPG methods on the competitive-cartpoles environment. MADDPG training represents a symmetric environment, and ST-MADDPG training gives an leader advantage to player~1. For each of these two methods, we created four pairs of agents with four different random seeds. In order to compare the agents' performance between the two training methods, we ran a tournament resulting in 320 game scores and trajectories for each of the methods. The tournament details are discussed in Appendix~\ref{sec:apdx_cartpoles}. The first two columns in Figure~\ref{fig:c_cartpole} summarize the statistics for the two tournaments. Note that the tournament game scores in this section refer to the scores of player~1. Therefore, a game will have a positive score if player~1 wins, a negative score if player~2 wins, and zero if the two players are tied.

Under the symmetric setting~(i.e.,~MADDPG), the performance of player~1 and player~2 are similar. The tournament has a mean score of $49.8$. While the majority of the games were scored between $-90$ to $90$, the rest of the games covered almost the entire score range from $-877$ to $920$. This indicates that while the two players have similar performance in most cases, each of them can occasionally outperform the other by a lot. In contrast, when given a leader advantage during training ~(i.e.,~ST-MADDPG), player~1 won more games with a larger mean score of $135$. Player~2 only got $-87$ on its best win, meaning that the follower could never significantly outperform the leader. Therefore, the leader has better overall performance compared to the follower. When observing the agents' behaviors by replaying the collected trajectories, we found that the two players resulting from the symmetric environment usually compete intensively by pushing and pulling each other via the spring. While they are able to keep their own poles upright, they fail to break the balance of the other agent and win the game in most of the competitions. Meanwhile, for the agents from ST-MADDPG, the leader manages to learn a policy to pull the follower out of the frame to win the game. Video demonstration of the robots' behaviors can be found in \href{https://sites.google.com/cs.washington.edu/stackelbergco-evolution/}{the project website}.

\paragraph{Re-balancing Asymmetric Environment.}
Given that ST-MADDPG creates an advantage that improves the leader's performance, we want to test if this can be used to compensate for a disadvantage that is assigned to an agent by the asymmetric environment. We created an asymmetric competitive-cartpoles environment by giving player~1 a force disadvantage, where player~1 has a decreased maximum control effort that is only $30\%$ as much as player~2's maximum effort. Afterward, we once again trained agents using both MADDPG and ST-MADDPG~(player~1 as the leader) with four random seeds and generated evaluation data with tournaments. As shown in Figure~\ref{fig:c_cartpole}, under a substantial force disadvantage, player~1's performance was significantly worse than player~2 after the MADDPG training. However, when Stackelberg gradient updates are applied, the two players' performances are equivalent. With a mean score of $4.57$, maximum score of $751$, and a minimum score of $-852$, \textbf{the leader advantage is able to compensate for the force disadvantage} for player~1 and generated a score distribution that is similar to the symmetric environment described above.

\subsection{Hopper Against Adversarial and Random Disturbance}
\label{sec:hopper_adv}
This work focuses on deriving and implementing a variation of MADDPG algorithm that uses the total derivative learning update to construct the Stackelberg information structure. While other individual learning algorithms are also popular in auto-curriculum works~\citep{dennis2020emergent, baker2019emergent}, their decentralized value networks estimation via surrogate functions makes the direct total derivative computation infeasible.
Alternatively, the Stackelberg information structure can be approximated by using different amounts of update steps for the leader and the follower in an individual learning setting~\citep{rajeswaran2020game}. The follower's best response update in a Stackelberg game is approximated by allowing the follower to make significantly more update steps than the leader for each batch of training data. 

In this robust control problem, we first compare the ST-MADDPG with MADDPG. We then also evaluated the approximated Stackelberg update with two PPO-based training settings. The first PPO approach trains all agents with the original PPO in a decentralized manner, which is commonly used in auto-curriculum literature~\citep{dennis2020emergent, baker2019emergent, yang2021motivating}. Afterward, we created a PPO variation with an approximated Stackelberg dynamic, ST-PPO, by having the follower~(adversarial disturbance) to take ten times more update steps than the leader~(hopper).

As shown in Table.\ref{tab:hopper}, the ST-MADDPG-trained hopper agents significantly outperformed the MADDPG-trained hopper agents under both adversarial attacks (ST-MADDPG: 5113.6 avg.~reward; MADDPG: 1601.6 avg.~reward; avg. \textbf{319.3\% improvement}) and random disturbances with multiple intensity levels (ST-MADDPG's average rewards were at least \textbf{2.22$\times$ better} than MADDPG). In contrast, the approximated Stackelberg information structure could only provide a mild (avg. \textbf{11\%}) improvement to the leader during adversarial training. Similarly, the hopper agents from ST-PPO were slightly better (at most \textbf{1.14$\times$}) than those from PPO in the random disturbance test. Therefore, \textbf{providing the leader advantage to the robot in adversarial training can further improve the robustness of a robot control policy.} With a similar computational complexity (detailed in \ref{sec:tds_vs_as}), \textbf{the total derivative update was more effective than the approximated Stackelberg update in terms of constructing the leader advantage.} Detail discussion on this experiment can found in Appendix~\ref{sec:apdx_hopper}.


\begin{figure*}[t!]
\centering
\includegraphics[width=0.96\textwidth]{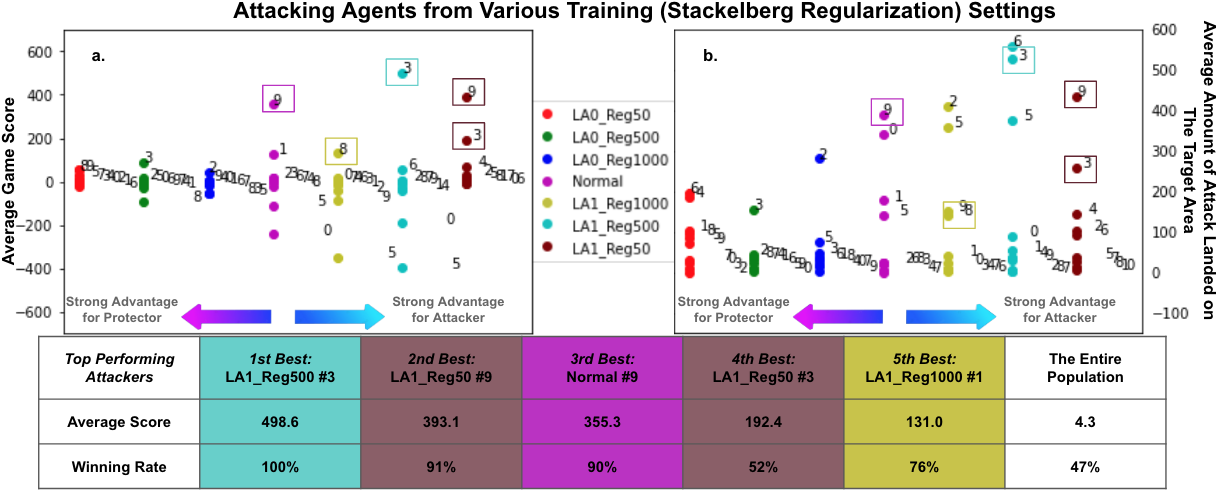}
\caption{The performance and behavior of attackers from seven training settings are visualized. The name of each group provides information about the training setting, where the first part indicates the leader agent(i.e.~LA0:Protector,~LA1:Attacker), and the second part indicates the regularization value. There are ten agents in each setting, the ID of each agent is printed on the \textbf{right side} of the corresponding point on both plots. The ID is located further right from the data point with a smaller corresponding Y value. \textbf{a.} Shows the average game score of 100 games between each attacker and a strong unseen baseline protector. \textbf{b.} Visualizes the average amount of attacks executed by an attacker in a single game~(i.e.~aggressiveness). The top five high performing agents (based on achieved average score) are highlighted in both plots.}
\label{fig:fencing}
\end{figure*}

\subsection{Co-evolution Under Complex Environment}
\label{sec:fencing}
The experiments above are based on two environments with simple robot dynamics and game rules, yet, practical robotic scenarios can be extensively more complicated. To provide more insight into competitive co-evolution in a complex environment under the Stackelberg structure, we evaluated various training settings on the Fencing Game. Compared to the previous environments, the Fencing Game represents more realistic challenges from practical robot problems. The complex kinematics of the two seven degree-of-freedom humanoid agents and the contact-rich nature of the game introduces large complexity and uncertainty to the environment transition model. Moreover, its game mechanism is also asymmetric. Since the protector is only rewarded by making contact with the attacker within the target area, the attacker leads the co-evolution process by initiating the attacking actions. If the attacker learns to stop attacking to avoid penalties, the policy updates for both agents could be less effective in that they become stuck in exploring highly sub-optimal areas of the task space. Therefore, the attacker plays a more important role in the game and we will focus on evaluating the attacker's behavior and performance in this experiment.

To study how various Stackelberg settings affect the quality of a co-evolution process, we compared the attackers from MADDPG and six different ST-MADDPG settings. For the six ST-MADDPG settings, we assigned the protector and the attacker as the leader respectively with small($\lambda$=50), medium($\lambda$=500), and large($\lambda$=1000) regularization values. The experiment tested the co-evolution process with each of the two agents having strong, medium, and weak leader advantages. Ten pairs of agents are trained with 10 different random seeds for each of the seven settings. We then evaluated each of the 70 trained attackers with 100 games against a carefully designed heuristic-based protector policy. By placing the protector's sword in between the target area and the point on the attacker's sword closest to the target area, the heuristic-based policy exploits embedded knowledge of the game's rules to execute a strong defensive strategy. Having this heuristic baseline policy allows fair comparison between attackers. Co-evolution processes that converge to a higher quality equilibrium should result in a more robust attacker strategy with better performance when competing against this strong unseen protector. The design of this heuristic-based policy, and the experimental details are described in Appendix~\ref{sec:apdx_fencinggame}.

\paragraph{Increase of Attackers' Aggressiveness.}
In all seven training settings, a large portion of the attackers converged to a set of conservative strategies that execute fewer attacking actions with relatively low positive(winning) or negative(losing) game scores. This result is not surprising given the fact that initiating an attack action also correlates to a huge risk of being penalized by the protector. However, as shown in Fig.\ref{fig:fencing}.b., when the attackers became more and more advantageous in the training, a small portion of the agents became increasingly aggressive in terms of landing more attacks on the target area.

\paragraph{Emergent Complexity.}
Increased engagement in the competition could help develop better attack strategies. Indeed, Fig.\ref{fig:fencing}.a. shows that the top five best-performing attackers are all located on the right half of the plot, where their co-evolving opponents are not the leader of a Stackelberg game. The top two attackers come from two ST-MADDPG settings where the attackers were the leader with a regularization value of 500 and 50 respectively. They both learned to trick the protector to move to a less manipulable joint configuration, and then attack the target area with low risk while the protector is partially trapped and busy moving out of that configuration. Of course, some other aggressive policies can also result in being overly engaged with poor performance~(e.g. agent\#5 from LA1\_Reg1000 and LA1\_Reg500 in Fig.\ref{fig:fencing}).

\section{Conclusion and Limitations}
This work studies the application of MARL to asymmetric, physically grounded competitive games. We proposed the Stackelberg-MADDPG algorithm, which formulates a two-player MARL problem as a Stackelberg game and provides an advantage to one of the agents in the system. We demonstrated that an agent's inherent advantage over the other could bias the training process towards unfavorable equilibria, and our proposed algorithm re-balances such environments in order to improve the quality of both agents. Our experiments also indicate that some problems can be more sensitive to random seeds, which could make the effect of the proposed method less obvious. Other limitations of this work include only being experimented with in two-player competitive settings and the need to access the opponent's parameters during training.



\clearpage


\bibliography{example}  

\begin{thebibliography}{26}
\providecommand{\natexlab}[1]{#1}
\providecommand{\url}[1]{\texttt{#1}}
\expandafter\ifx\csname urlstyle\endcsname\relax
  \providecommand{\doi}[1]{doi: #1}\else
  \providecommand{\doi}{doi: \begingroup \urlstyle{rm}\Url}\fi

\bibitem[Zhang et~al.(2019)Zhang, Yang, and Ba{\c{s}}ar]{zhang2019multi}
K.~Zhang, Z.~Yang, and T.~Ba{\c{s}}ar.
\newblock Multi-agent reinforcement learning: A selective overview of theories
  and algorithms.
\newblock \emph{arXiv preprint arXiv:1911.10635}, 2019.

\bibitem[Filar and Vrieze(2012)]{filar2012competitive}
J.~Filar and K.~Vrieze.
\newblock \emph{Competitive Markov decision processes}.
\newblock Springer Science \& Business Media, 2012.

\bibitem[Silver et~al.(2017)Silver, Hubert, Schrittwieser, Antonoglou, Lai,
  Guez, Lanctot, Sifre, Kumaran, Graepel, et~al.]{silver2017mastering}
D.~Silver, T.~Hubert, J.~Schrittwieser, I.~Antonoglou, M.~Lai, A.~Guez,
  M.~Lanctot, L.~Sifre, D.~Kumaran, T.~Graepel, et~al.
\newblock Mastering chess and shogi by self-play with a general reinforcement
  learning algorithm.
\newblock \emph{arXiv preprint arXiv:1712.01815}, 2017.

\bibitem[Berner et~al.(2019)Berner, Brockman, Chan, Cheung, P.~Debiak,
  Dennison, Farhi, Fischer, Hashme, Hesse, et~al.]{berner2019dota}
C.~Berner, G.~Brockman, B.~Chan, V.~Cheung, P.~P.~Debiak, C.~Dennison,
  D.~Farhi, Q.~Fischer, S.~Hashme, C.~Hesse, et~al.
\newblock Dota 2 with large scale deep reinforcement learning.
\newblock \emph{arXiv preprint arXiv:1912.06680}, 2019.

\bibitem[Vinyals et~al.(2019)Vinyals, Babuschkin, Czarnecki, Mathieu, Dudzik,
  Chung, Choi, Powell, Ewalds, Georgiev, et~al.]{vinyals2019grandmaster}
O.~Vinyals, I.~Babuschkin, W.~M. Czarnecki, M.~Mathieu, A.~Dudzik, J.~Chung,
  D.~H. Choi, R.~Powell, T.~Ewalds, P.~Georgiev, et~al.
\newblock Grandmaster level in starcraft ii using multi-agent reinforcement
  learning.
\newblock \emph{Nature}, 575\penalty0 (7782):\penalty0 350--354, 2019.

\bibitem[Dennis et~al.(2020)Dennis, Jaques, Vinitsky, Bayen, Russell, Critch,
  and Levine]{dennis2020emergent}
M.~Dennis, N.~Jaques, E.~Vinitsky, A.~Bayen, S.~Russell, A.~Critch, and
  S.~Levine.
\newblock Emergent complexity and zero-shot transfer via unsupervised
  environment design.
\newblock \emph{arXiv preprint arXiv:2012.02096}, 2020.

\bibitem[Baker et~al.(2019)Baker, Kanitscheider, Markov, Wu, Powell, McGrew,
  and Mordatch]{baker2019emergent}
B.~Baker, I.~Kanitscheider, T.~Markov, Y.~Wu, G.~Powell, B.~McGrew, and
  I.~Mordatch.
\newblock Emergent tool use from multi-agent autocurricula.
\newblock \emph{arXiv preprint arXiv:1909.07528}, 2019.

\bibitem[Yang et~al.(2021)Yang, Habibi, Lancaster, Boots, and
  Smith]{yang2021motivating}
B.~Yang, G.~Habibi, P.~Lancaster, B.~Boots, and J.~Smith.
\newblock Motivating physical activity via competitive human-robot interaction.
\newblock In \emph{5th Annual Conference on Robot Learning}, 2021.

\bibitem[Pinto et~al.(2017)Pinto, Davidson, Sukthankar, and
  Gupta]{pinto2017robust}
L.~Pinto, J.~Davidson, R.~Sukthankar, and A.~Gupta.
\newblock Robust adversarial reinforcement learning.
\newblock In \emph{International Conference on Machine Learning}, pages
  2817--2826. PMLR, 2017.

\bibitem[Bansal et~al.(2017)Bansal, Pachocki, Sidor, Sutskever, and
  Mordatch]{bansal2017emergent}
T.~Bansal, J.~Pachocki, S.~Sidor, I.~Sutskever, and I.~Mordatch.
\newblock Emergent complexity via multi-agent competition.
\newblock \emph{arXiv preprint arXiv:1710.03748}, 2017.

\bibitem[Won et~al.(2021)Won, Gopinath, and Hodgins]{won2021control}
J.~Won, D.~Gopinath, and J.~Hodgins.
\newblock Control strategies for physically simulated characters performing
  two-player competitive sports.
\newblock \emph{ACM Transactions on Graphics (TOG)}, 40\penalty0 (4):\penalty0
  1--11, 2021.

\bibitem[Foerster et~al.(2018)Foerster, Chen, Al-Shedivat, Whiteson, Abbeel,
  and Mordatch]{foerster2017learning}
J.~Foerster, R.~Y. Chen, M.~Al-Shedivat, S.~Whiteson, P.~Abbeel, and
  I.~Mordatch.
\newblock Learning with opponent-learning awareness.
\newblock In \emph{Proceedings of the 17th International Conference on
  Autonomous Agents and MultiAgent Systems}, AAMAS '18, page 122–130, 2018.

\bibitem[Prajapat et~al.(2020)Prajapat, Azizzadenesheli, Liniger, Yue, and
  Anandkumar]{prajapat2020competitive}
M.~Prajapat, K.~Azizzadenesheli, A.~Liniger, Y.~Yue, and A.~Anandkumar.
\newblock Competitive policy optimization.
\newblock \emph{arXiv preprint arXiv:2006.10611}, 2020.

\bibitem[Zheng et~al.(2021)Zheng, Fiez, Alumbaugh, Chasnov, and
  Ratliff]{zheng2021stackelberg}
L.~Zheng, T.~Fiez, Z.~Alumbaugh, B.~Chasnov, and L.~J. Ratliff.
\newblock Stackelberg actor-critic: Game-theoretic reinforcement learning
  algorithms.
\newblock \emph{arXiv preprint arXiv:2109.12286}, 2021.

\bibitem[Lin et~al.(2020)Lin, Jin, and Jordan]{lin2020near}
T.~Lin, C.~Jin, and M.~I. Jordan.
\newblock Near-optimal algorithms for minimax optimization.
\newblock In \emph{Conference on Learning Theory}, pages 2738--2779. PMLR,
  2020.

\bibitem[Fiez et~al.(2020)Fiez, Chasnov, and Ratliff]{fiez2020implicit}
T.~Fiez, B.~Chasnov, and L.~J. Ratliff.
\newblock Implicit learning dynamics in stackelberg games: Equilibria
  characterization, convergence analysis, and empirical study.
\newblock In \emph{International Conference on Machine Learning}, 2020.

\bibitem[Von~Stackelberg(2010)]{von2010market}
H.~Von~Stackelberg.
\newblock \emph{Market structure and equilibrium}.
\newblock Springer Science \& Business Media, 2010.

\bibitem[Yang and Wang(2020)]{yang2020overview}
Y.~Yang and J.~Wang.
\newblock An overview of multi-agent reinforcement learning from game
  theoretical perspective.
\newblock \emph{arXiv preprint arXiv:2011.00583}, 2020.

\bibitem[Ba{\c{s}}ar and Olsder(1998)]{bacsar1998dynamic}
T.~Ba{\c{s}}ar and G.~J. Olsder.
\newblock \emph{Dynamic noncooperative game theory}.
\newblock SIAM, 1998.

\bibitem[Lowe et~al.(2017)Lowe, WU, Tamar, Harb, Pieter~Abbeel, and
  Mordatch]{lowe2017multi}
R.~Lowe, Y.~WU, A.~Tamar, J.~Harb, O.~Pieter~Abbeel, and I.~Mordatch.
\newblock Multi-agent actor-critic for mixed cooperative-competitive
  environments.
\newblock \emph{Advances in Neural Information Processing Systems},
  30:\penalty0 6379--6390, 2017.

\bibitem[Krantz and Parks(2002)]{krantz2002implicit}
S.~G. Krantz and H.~R. Parks.
\newblock \emph{The implicit function theorem: history, theory, and
  applications}.
\newblock Springer Science \& Business Media, 2002.

\bibitem[Duan et~al.(2016)Duan, Chen, Houthooft, Schulman, and
  Abbeel]{duan2016benchmarking}
Y.~Duan, X.~Chen, R.~Houthooft, J.~Schulman, and P.~Abbeel.
\newblock Benchmarking deep reinforcement learning for continuous control.
\newblock In \emph{International conference on machine learning}, pages
  1329--1338. PMLR, 2016.

\bibitem[Rajeswaran et~al.(2020)Rajeswaran, Mordatch, and
  Kumar]{rajeswaran2020game}
A.~Rajeswaran, I.~Mordatch, and V.~Kumar.
\newblock A game theoretic framework for model based reinforcement learning.
\newblock In \emph{International conference on machine learning}, pages
  7953--7963. PMLR, 2020.

\bibitem[Pearlmutter(1994)]{pearlmutter1994fast}
B.~A. Pearlmutter.
\newblock Fast exact multiplication by the hessian.
\newblock \emph{Neural computation}, 6\penalty0 (1):\penalty0 147--160, 1994.

\bibitem[Rajeswaran et~al.(2019)Rajeswaran, Finn, Kakade, and
  Levine]{rajeswaran2019meta}
A.~Rajeswaran, C.~Finn, S.~M. Kakade, and S.~Levine.
\newblock Meta-learning with implicit gradients.
\newblock \emph{Advances in neural information processing systems}, 32, 2019.

\bibitem[Lillicrap et~al.(2015)Lillicrap, Hunt, Pritzel, Heess, Erez, Tassa,
  Silver, and Wierstra]{lillicrap2015continuous}
T.~P. Lillicrap, J.~J. Hunt, A.~Pritzel, N.~Heess, T.~Erez, Y.~Tassa,
  D.~Silver, and D.~Wierstra.
\newblock Continuous control with deep reinforcement learning.
\newblock \emph{arXiv preprint arXiv:1509.02971}, 2015.

\end{thebibliography}

\newpage
\appendix
\section{Algorithm and Experiment Details}
\subsection{ST-MADDPG}
\label{sec:apdx_stmaddpg}
Algorithm~\ref{comp:alg:stmaddpg} details the ST-MADDPG algorithm proposed in this paper. In this work we select the implicit map regularization hyperparameter $\lambda$ via a grid search for the first and second environment. In general, the neural network may be highly non-convex and the hessian inverse can be ill-conditioned. A larger regularization prevents the gradient from exploding and yields smoother learning dynamics, as observed in our experiments as well as in other Stackelberg learning applications~\citep{fiez2020implicit, zheng2021stackelberg}. How to trade-off between Stackelberg and normal gradient learning by picking the regularization optimally or even adaptively is a future direction.

\begin{algorithm}[h!]
\caption{ST-MADDPG algorithm}
\SetAlgoLined
	\For{$\mathrm{episodes}$ $k=1, 2, \dots, K$}{
		receive initial state $s_0$\;
		\For{$t = 1, 2, \dots, T$ }{
			for each agent $i$, select action $a^i = \mu_{\theta_i}(s)$ according to the corrent policy\;
			execute actions $(a^1, a^2)$ and observe reward $r$ and new state $s'$\;
			store $(s, a^1, a^2, r, s')$ in replay buffer $\mathcal{D}$\;
			$s \leftarrow s'$\;
			sample a random minibatch of $N$ transitions $(s_j, a_j^1, a_j^2, r_j, s_j')$ from $\mathcal{D}$\;
			set $y_j = r_j + Q_{w'}(s_j, \mu_{\theta_1'}(s_j), \mu_{\theta_2'}(s_j))$\;
			update the critic by minimizing the loss: $$L(w) =  \frac{1}{N} \sum_{i=1}^N [ \left(Q_w(s_j, a_j^1, a_j^2) - y_j \right)^2 ]$$
			
			update the leader policy using the total gradient computed by~\eqref{comp:eqn:total_deri}:
			$$\theta_1 \leftarrow \theta_1 + \alpha^1 \nabla^\lambda J(\theta_1, \theta_2)$$
			
			update the follower policy using the policy gradient:
			$$\theta_2 \leftarrow \theta_2 - \alpha^2 \nabla_{\theta_2} J(\theta_1, \theta_2)$$
			
			update the target networks:
			$$w' \leftarrow \tau w + (1-\tau) w'$$
			$${\theta_i}' \leftarrow \tau \theta_i + (1-\tau) {\theta_i}'$$
		}
	}
	\label{comp:alg:stmaddpg}
\end{algorithm}

\subsubsection{Computational Complexity}
\label{sec:stmaddpg_comp}
Unlike the standard policy gradient update, the leader's policy update requires the computation of a Jacobian-vector product for the $\nabla_{\theta_1\theta_2}J(\theta_1, \theta_2)$ term and an inverse-Hessian-vector product, $\nabla^2_{\theta_2}J(\theta_1, \theta_2)\nabla_{\theta_2}J(\theta_1, \theta_2)$, in equation \ref{comp:eqn:total_deri}. The automatic differentiation engine in PyTorch calculates the Jacobian-vector product. The inverse-Hessian-vector term is calculated by the iterative conjugate gradient (CG) method that we also implemented with the automatic differentiation engine. Therefore, all the calculations for the Stackelberg gradient update step happened on GPU. Each iteration of CG calculates a Hessian vector product that takes $\sim1.5$ times the cost of a gradient~\citep{pearlmutter1994fast}. Empirically, five iterations of CG can sufficiently achieve numerical precision, so the leader update costs $\sim7.5$ times a normal gradient~\citep{rajeswaran2019meta, fiez2020implicit}. However, in practice, ST-MADDPG only needs approximately twice the time of MADDPG because the main bottleneck of reinforcement learning is in trajectory sampling rather than gradient calculation.

\subsection{Competitive-Cartpoles}
\label{sec:apdx_cartpoles}
The maximum length of the competitive-cartpoles game is 1000 time steps in training and experiments. At the beginning of the training, each agent runs $10^4$ steps of uniform-random action selection for better exploration \cite{lillicrap2015continuous}. The reply buffer size was set to $10^6$, with a learning rate of $10^{-3}$, and each agent was updated with a batch size of $10^2$. We ran two tournaments to sample evaluation data, one for the agents resulting from MADDPG and the other for those from ST-MADDPG. In a tournament, each of the four player~1 agents~(resulted from four random seeds) played 20 games against each of the four player~2 agents, resulting in 320 game scores and trajectories. The regularization values $\lambda$ in all ST-MADDPG training were set to one, such that it prevents the gradient from exploding while still creating a very strong leader advantage in the training. This evaluation process allows an agent to play games with not only its original co-evolving opponent but also the opponents that are trained in different random seeds, providing a more comprehensive summary for each training set.

\subsection{Hopper}
\label{sec:apdx_hopper}
During training, the maximum length of the games was bounded by 1000 time steps for a shorter training time. However, in the evaluation experiments, all trials have a maximum length of 3000 time steps to better distinguish agents' performance. Similar to the competitive-cartpole environment, each training process in this environment begins with $10^4$ steps of uniform-random action selection for better exploration \cite{lillicrap2015continuous}. The reply buffer size was set to $10^6$, with a learning rate of $8 \times 10^{-5}$, and each agent was updated with a batch size of $10^2$. In this environment, the regularization values $\lambda$ in all ST-MADDPG training were set to 5000.



\subsubsection{Experimental Result}
We used MADDPG, ST-MADDPG, PPO, ST-PPO algorithms to create 20 pairs of hoppers and adversaries with 20 random seeds respectively. Each pair of agents were evaluated with 10 games, resulting in 200 game scores for each method.
\paragraph{Adversarial Attack.} We found that the ST-MADDPG trained hoppers were able to survive significantly longer~(\textbf{5113.6} avg.~reward) than those from MADDPG training~(\textbf{3333.0} avg.~reward) under adversarial attacks. However, the hopper agents from ST-PPO~(\textbf{522.0} avg.~reward) survived marginally longer than those from PPO~(\textbf{473.6} avg.~reward).

\paragraph{High Intensity Random Disturbance.} The generalizability of all hopper agents was tested under three environments with no disturbances and two different levels of strong random disturbances respectively~(e.g.,~$0N$, $0.1N$, and $10.N$). Each agent ran 100 trials in each of the environments, and Table.\ref{tab:hopper} compares the agents' performance from the two training methods. Even though the maximum strength of the adversaries in training was bounded by $0.001N$, some of the agents~(e.g.,~MADDPG:$50\%$, ST-MADDPG:$100\%$) still managed to receive more than 1000 average rewards under random disturbances with the maximum strength of $10N$. ST-MADDPG trained agents greatly outperformed MADDPG trained agents in all three levels of intensities. Therefore, ST-MADDPG policies are robust enough to maintain high performance under unseen scenarios. 

\subsubsection{Total Derivative Stackelberg Update V.S. Approximated Stackelberg Update}
\label{sec:tds_vs_as}
As discussed in section~\ref{sec:stmaddpg_comp}, the total derivative Stackelberg update step takes $\sim7.5$ times a normal gradient step. In our experiment, the approximated Stackelberg update for the follower takes ten times more update steps than the leader. For example, if a regular policy gradient method required time $t$ to update the policy for a single agent in each epoch (trajectory sampling time not included), the total derivative update method takes $7.5t$ for the leader update, and $t$ for the follower update. And the approximation update method takes $t$ for the leader update, and $10t$ for the follower update. Therefore, the computational complexity of the total derivative update is slightly smaller than the approximation update in our experiment setting. With a similar computation complexity, the total derivative update was significantly more effective on constructing the Stackelberg information structure. The effectiveness of the approximation approach can be improved by further increasing the amount of the follower's update steps with computational trade-off. As observed in \citet{rajeswaran2020game}, the approximation method's performance is more acceptable when the follower takes $25\times$ more update steps then the leader.

\begin{table}[!h]
\begin{center}
\begin{tabular}{ |c||c|c|c|c| } 
\hline
\multicolumn{5}{| c |}{Adversarial Disturbance}\\
\hline
     & ST-MADDPG & MADDPG & ST-PPO & PPO\\
\hline
Avg. Score& 5113.6 & 1601.6 & 521.8 & 473.6\\ 
Std       & 3333.0 & 753.2  & 147.1 & 91.4\\
\hline
Statistical Significance&\multicolumn{2}{| c |}{$p < 0.01$}&\multicolumn{2}{| c |}{$p > 0.05$} \\
\hline
\hline
\multicolumn{5}{| c |}{No Random Disturbance}\\
\hline
     & ST-MADDPG & MADDPG & ST-PPO & PPO\\
\hline
Avg. Score& 5415.2 & 2436.8 & 510.0 & 495.7 \\ 
Std       & 3730.8 & 2265.5 & 84.6  & 102.0\\
\hline
Statistical Significance&\multicolumn{2}{| c |}{$p < 0.05$}&\multicolumn{2}{| c |}{$p > 0.05$} \\
\hline
\hline
\multicolumn{5}{| c |}{0.1N Random Disturbance}\\
\hline
     & ST-MADDPG & MADDPG & ST-PPO & PPO\\
\hline
Avg. Score& 5463.3 & 2423.6 & 530.9 & 486.0\\ 
Std       & 3575.6 & 2178.2 & 101.3 & 75.7\\
\hline
Statistical Significance&\multicolumn{2}{| c |}{$p < 0.05$}&\multicolumn{2}{| c |}{$p > 0.05$} \\
\hline
\hline
\multicolumn{5}{| c |}{10 N Random Disturbance}\\
\hline
     & ST-MADDPG & MADDPG & ST-PPO & PPO\\
\hline
Avg. Score& 4928.3 & 1991.2 & 540.1 & 474.3\\ 
Std       & 3588.0 & 2084.6 & 151.3 & 80.8\\
\hline
Statistical Significance&\multicolumn{2}{| c |}{$p < 0.05$}&\multicolumn{2}{| c |}{$p > 0.05$} \\
\hline
\end{tabular}
\caption{Performance comparison between ST-MADDPG, MADDPG, ST-PPO, and PPO trained Hopper agents under three levels of random disturbances. Evaluation data between the Stackelberg and regular versions of both algorithms are compared via a statistic significant test (i.e. u-test)}
\label{tab:hopper}
\end{center}
\end{table}

\subsection{The Fencing Game}
\label{sec:apdx_fencinggame}
\paragraph{Game Rules.} Algorithm~\ref{alg:game-score} summarizes the scoring mechanism of the fencing game. The attacker (or antagonist on fig.\ref{fig:all_env}) will get one point by placing its sword within the orange spherical(target) area located between the two agents. But the attacker will receive a negative ten points of score penalty if its sword is placed within the target area and makes contact with the protector's (or protagonist on fig.\ref{fig:all_env}) sword simultaneously. Meanwhile, the goal for the protector agent on the left is to minimize the attacker's score by giving him score penalties. Additionally, the attacker will get 10 points of reward if the protector's sword is placed within the target area, passively waiting for the attacker to attack for more than 2 seconds. Each agent has a seven dimension continuous control space. Each game will last for 1000 time-steps~(i.e.~10 seconds)

\begin{algorithm}[H]
        \SetAlgoLined
    \textbf{Initialize:} Game score \textit{s} = 0; Timestep = 0.01 Sec; Game horizon = 10 Sec \\
     bat\_a $\rightarrow$ Antagonist's bat\\
     bat\_p $\rightarrow$ Protagonist's bat\\
     target $\rightarrow$ Target Area\\
     \For{every timestep in this game}{
     \uIf{bat\_a in target}{
        \eIf{bat\_a contacts bat\_p}
        {\textit{s} -= 10}
        {\textit{s} += 1}
      }
      
     \uIf{bat\_p in target for more than 200 consecutive timesteps}
      {\textit{s} += 10}
     }
\caption{The Fencing Game Scoring Mechanism}
\label{alg:game-score}
\end{algorithm}

\paragraph{Heuristic-based Protagonist Policy.} We aimed to design a strong baseline heuristic policy to create an intense human robot gameplay experience. Given an observation of the world, the robot orients its bat perpendicular to the human's bat with random angular offsets drawn uniformly from -25 to 25 degrees on the x, y, and z axes. In order to ensure that the robot is always executing a competitive defense, the policy commands the robot to position the center of its bat in between the target area and the point on the human's bat that is closest to the target area:
\begin{gather*}
\bar{b_p} = \bar{tar} + (\bar{h_{close}} - \bar{tar}) \cdot uniform(0.5, 1) \\
\bar{h_{close}} = \bar{h_{low}} + ht \cdot (\bar{h_{up}} - \bar{h_{low}}) \\
ht = \max\big( 0, \min\big(1, (\bar{tar} - \bar{h_{low}}) \cdot (\bar{h_{up}} - \bar{h_{low}}) / (2\cdot L_{sword}) \big)  \big)
\end{gather*}
Where $\bar{b_p}$, $\bar{tar}$, $\bar{h_{up}}$ and $\bar{h_{low}}$ represent the position of the robot's bat frame, the center of the target area, the upper end of human's bat, and the lower end of human's bat respectively. $\bar{h_{close}}$ indicates the point on the human's bat that is closest to the center of the target area, and $L_{sword}$ indicates the length of a bat. The function $uniform(0.5, 1)$ randomly determines how far apart the robot's bat should be from the human's bat. In addition, there is a 50\% chance for the robot to execute the desired bat position calculated from the last time step instead of the latest desired pose. The added uncertainties introduce randomness to the robot's behavior. This heuristic allows the robot to dominate the fencing game when it can move faster or as fast as the antagonist. In this experiment, the physical capability of the antagonist and protagonist agents are identical.

\paragraph{Pre-training.}
Because the fencing environment is significantly more complex compare to the other two experimental environments. We choose to initialize the co-evolution process of all seven setting (i.e. LA0\_Reg50, LA0\_Reg500, LA0\_Reg1000, Normal, LA1\_Reg50, LA1\_Reg500, and LA1\_Reg1000 in Fig.\ref{fig:fencing}) with the same pair of pre-trained policies. The pre-trained policies were created by iteratively running two rounds of best response individual update for each agent using MADDPG, which is similar to the pre-training process of \citet{yang2021motivating}. Each iteraction of best respond update starts with $10^4$ steps of uniform-random action selection. The reply buffer size was set to $10^6$, with a learning rate of $8 \times 10^{-4}$, and each agent was updated with a batch size of $100$.

\begin{figure*}[t!]
\centering
\includegraphics[width=0.85\textwidth]{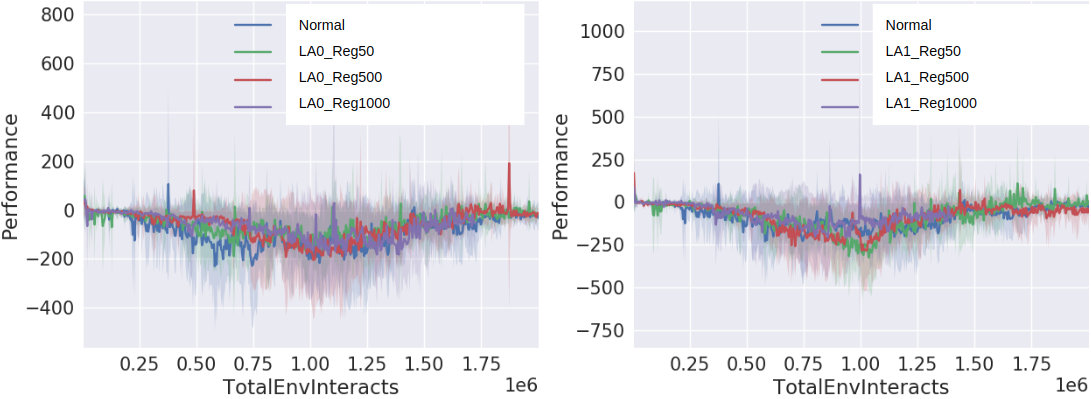}
\caption{Both plots demonstrate the protector's (i.e.~agent 0) average game score during the co-evolution process under the six different settings.}
\label{fig:twolr}
\end{figure*}

\paragraph{Co-evolution.}
The protector(i.e.~agent 0) and attacker(i.e.~agent 1) agents were initialized by the pre-trained policies for all seven training settings under all 10 random seeds. There was no uniform-random action selection at the beginning of each co-evolution process. The reply buffer size was set to $10^6$, with a learning rate of $8 \times 10^{-5}$, and each agent was updated with a batch size of $1024$. As Fig.\ref{fig:twolr} shows, the agents diverged out of the pre-trained policies and had drastic policy updates with signicicant performance fluctuation in the middle of the co-evolution process. The agents then converged to a new equilibrium by the end of the co-evolution process.

$\pi^2_{\theta_2}$

\end{document}